\documentclass{article}

\usepackage{microtype}
\usepackage{graphicx}
\usepackage{subfigure}
\usepackage{booktabs} % for professional tables
\usepackage{url}
\usepackage{amsmath,amsthm,amssymb,amsfonts} 
\usepackage{nicefrac}
\usepackage{mathtools}
\usepackage{balance}
\mathtoolsset{showonlyrefs}
\usepackage{enumitem}
\usepackage{wrapfig}
\usepackage{multirow}
\usepackage{hhline}

\usepackage{hyperref}

\usepackage[accepted]{icml2021}

\renewcommand{\Pr}{\field{P}}

\newcommand{\bg}{\boldsymbol{g}}
\newcommand{\bx}{\boldsymbol{x}}

\newcommand{\by}{\boldsymbol{y}}

\newcommand{\field}[1]{\mathbb{#1}}

\newcommand{\R}{\field{R}}

\newcommand{\E}{\field{E}}

\newtheorem{lemma}{Lemma}
\newtheorem{theorem}{Theorem}

\icmltitlerunning{A Second look at Exponential and Cosine Step Sizes}

\begin{document}
	


\twocolumn[

\icmltitle{A Second look at Exponential and Cosine Step Sizes:\\ Simplicity, Adaptivity, and Performance}

\begin{icmlauthorlist}
\icmlauthor{Xiaoyu Li $^*$}{se}
\icmlauthor{Zhenxun Zhuang $^*$}{cs}
\icmlauthor{Francesco Orabona}{se,cs,ee}
\end{icmlauthorlist}

\icmlaffiliation{se}{Division of System Engineering, Boston University, Boston, MA, US}
\icmlaffiliation{cs}{Department of Computer Science, Boston University, Boston, MA, US}
\icmlaffiliation{ee}{Department of Electrical \& Computer Engineering, Boston University, Boston, MA, US}

\icmlcorrespondingauthor{Xiaoyu Li}{xiaoyuli@bu.edu}
\icmlcorrespondingauthor{Zhenxun Zhuang}{zxzhuang@bu.edu}

\icmlkeywords{Stochastic Optimization, Non-convex, ICML}

\vskip 0.3in
]



\printAffiliationsAndNotice{\icmlEqualContribution} % otherwise use the standard text.
\begin{abstract}
Stochastic Gradient Descent (SGD) is a popular tool in training large-scale machine learning models. Its performance, however, is highly variable, depending crucially on the choice of the step sizes. Accordingly, a variety of strategies for tuning the step sizes have been proposed, ranging from coordinate-wise approaches (a.k.a. ``adaptive'' step sizes) to sophisticated heuristics to change the step size in each iteration.
In this paper, we study two step size schedules whose power has been repeatedly confirmed in practice: the exponential and the cosine step sizes. For the first time, we provide theoretical support for them proving convergence rates for smooth non-convex functions, with and without the Polyak-\L{}ojasiewicz (PL) condition. Moreover, we show the surprising property that these two strategies are \emph{adaptive} to the noise level in the stochastic gradients of PL functions. That is, contrary to polynomial step sizes, they achieve almost optimal performance without needing to know the noise level nor tuning their hyperparameters based on it. Finally, we conduct a fair and comprehensive empirical evaluation of real-world datasets with deep learning architectures. Results show that, even if only requiring at most two hyperparameters to tune, these two strategies best or match the performance of various finely-tuned state-of-the-art strategies.
\end{abstract}

\section{Introduction}
In the last 10 years, non-convex machine learning formulations have received more and more attention as they can typically better scale with the complexity of the predictors and the amount of training data compared with convex ones. One such example is the deep neural networks. Over the years, various algorithms have been proposed and employed to optimize non-convex machine learning problems, among which Stochastic Gradient Descent (SGD)~\citep{RobbinsM51} has become the most important ingredient in Machine Learning pipelines. Practitioners prefer it over more sophisticated methods for its simplicity and speed.
%that allow the same optimization algorithm to seamlessly be used from convex to non-convex domains.
Yet, this generality comes with a cost: SGD is far from the robustness of, e.g., second-order methods that require little to no tweaking of knobs to work. In particular, the step size is still the most important parameter to tune in the SGD algorithm, carrying the actual weight of making SGD adaptive to different situations. 

The importance of step sizes in SGD is testified by the numerous proposed strategies to tune step sizes~\citep[e.g.,][]{DuchiHS10, McMahanS10, TielemanH12, Zeiler12, KingmaB15}. However, for most of them, there is little or no theory that can really explain their empirical success. 
Moreover, SGD with appropriate step sizes is already optimal in all the possible situations, so it is unclear what kind of advantage we might show.

An interesting viewpoint is to go beyond worst-case analyses and show that these learning rates provide SGD with some form of \emph{adaptivity} to the characteristics of the function. More specifically, an algorithm is considered adaptive (or \emph{universal}) if it has the best theoretical performance w.r.t. to a quantity X without the need to know it~\citep{Nesterov15b}. So, for example, it is possible to design optimization algorithms adaptive to scale~\citep{OrabonaP15,OrabonaP18}, smoothness~\citep{LevyYC18}, noise~\citep{LevyYC18,LiO19}, and strong convexity~\citep{CutkoskyO18}.
On the other hand, as noted in \citet{Orabona19}, it is remarkable that even if most of the proposed step size strategies for SGD are called ``adaptive'', for most of them their analyses do not show any provable advantage over plain SGD nor any form of adaptation to the intrinsic characteristics of the non-convex function.
%Indeed, the word ``adaptive'' that has a precise meaning in statistics and machine learning, has currently assumed the meaning of ``coordinate-wise''.
%A notable exception is Adagrad~\citep{DuchiHS10, McMahanS10} that, under different non-convex assumptions, has been proven to be adaptive to noise in the stochastic gradients~\citep{LiO19} and robust to the mispecifications of some of its parameters~\citep{WardWB20}.

In this paper, we look at the two simple to use and empirically successful step size decay strategies, the \textit{exponential} and the \emph{cosine step size} (with and without restarts)~\citep{LoshchilovH17, HeZZZXL19}.
The exponential step size is simply an exponential decaying step size.
%and it requires only two hyperparameters.
It is less discussed in the optimization literature and it is also unclear who proposed it first, even if it has been known to practitioners for a long time and already included in many deep learning software libraries~\citep[e.g.,][]{Tensorflow15,Pytorch19}.
The cosine step size, which anneals the step size following a cosine function, has exhibited great power in practice but it does not have any theoretical justification.

For both these step size decay strategies, we prove \emph{for the first time} a convergence guarantee. Moreover, we show that they have (unsuspected!) adaptation properties.
% One of the most successful step strategies is the \emph{stagewise step decay}~\citep{KrizhevskySH12, SimonyanZ15, HeZRS16, HuangLVW17}.
% %Nevertheless, there has been a gap between theory and practice. Specifically, most of those proposed strategies are not justified by theory; while for those theoretically supported ones, their empirical performance cannot match the state of the art. One potential strategy that has merits of both sides is the \emph{stagewise step decay} whose empirical power has already been widely acknowledged.~\citep{KrizhevskySH12, SimonyanZ15, HeZRS16, HuangLVW17}.
% It starts the training with a relatively large constant step size, and then periodically decreases the step size, for example when the curve of the validation loss plateaus. Indeed, assuming we were able to know in advance some unknown quantities of the function to be optimized, it can also be shown that it would be theoretically optimal~\citep{HazanK11}. However, in practice, we typically do not have access to those quantities about the function so the tuning can be hard.
%Here, we equip the exponential step sizes and cosine step size with the theoretical support and we also underline situations in which they have a provable advantage over other schemes.
Moreover, we also empirically test them showing that they have the best empirical performance among various state-of-the-art strategies. Finally, our proofs reveal the hidden similarity between these two step sizes.

Specifically, the contributions of this paper are:
\begin{itemize}%[topsep=0pt]
\item In the case when the function satisfies the PL condition~\citep{Polyak63,Lojasiewicz63,KarimiNS16}, both exponential step size and cosine step size strategies \textit{automatically adapt to the level of noise of the stochastic gradients}. 
%Moreover, the rates obtained for the exponential step size are \emph{new} in the literature on optimization of smooth PL functions.
\item Without the PL condition, we show that SGD with either exponential step sizes or cosine step sizes has an \textit{(almost) optimal convergence rate} for smooth non-convex functions.
\item We also conduct an empirical evaluation on deep learning architectures: Exponential and cosine step sizes have essentially matching or better empirical performance than polynomial step decay, stagewise step decay, Adam~\citep{KingmaB15}, and stochastic line search~\citep{VaswaniMLSGLJ19}, while requiring at most two hyperparameters.
\end{itemize}

The rest of the paper is organized as follows: We first discuss the relevant literature (Section~\ref{sec:rel}). In Section~\ref{sec:def}, we introduce the notation, setting, and precise assumptions. Then, in Section~\ref{sec:theorem} we describe in detail the step sizes and the theoretical guarantees. We show our empirical results in Section~\ref{sec:exp}. Finally, we conclude with a discussion of the results and future work.

\section{Related Work}
\label{sec:rel}

\textbf{Adaptation in non-convex optimization} Adaptation is a general concept and an algorithm can be adaptive to any characteristic of the optimization problem. The idea is formalized in \citep{Nesterov15b} with the equivalent name of \emph{universality}, but it goes back at least to the ``self-confident'' strategies in online convex optimization~\citep{AuerCG02}. Indeed, the famous AdaGrad algorithm~\citep{McMahanS10,DuchiHS10} uses exactly that method to design an algorithm \emph{adaptive to the gradients}. Nowadays, ``adaptive step size'' tend to denote coordinate-wise ones, with no guarantee of adaptation to any particular property. There is an abundance of adaptive optimization algorithm in the convex setting~\citep[e.g.,][]{McMahanS10,DuchiHS10,KingmaB15,ReddiKK18}, while only a few in the more challenging non-convex setting~\citep[e.g.,][]{ChenZTYG18}. The first analysis to show adaptivity to noise of non-convex SGD with appropriate step sizes is in \citet{LiO19} and later in \citet{WardWB19,WardWB20} under stronger assumptions. Then, \citet{LiO20} studied the adaptivity to noise of AdaGrad plus momentum, with a high probability analysis.

\textbf{Exponential step size} To the best of our knowledge, the exponential step size has been incorporated in Tensorflow~\citep{Tensorflow15} and PyTorch~\citep{Pytorch19}, yet no convergence guarantee have ever been proved for it.
%\footnote{Despite the name, the exponential step size has no relationship with the one in~\citet{LiA20}, where an exponentially \emph{increasing} step size is analyzed under very special conditions.}
The closest strategy is the \emph{stagewise step decay}, which corresponds to the discrete version of the exponential step size we analyze.
The stagewise step decay uses a piece-wise constant step size strategy, where the step size is cut by a factor in each ``stage''.
This strategy is known with many different names: ``stagewise step size''~\citep{YuanYJY19}, ``step decay schedule''~\citep{GeKKN19}, ``geometrically decaying schedule''~\citep{DavisDXZ19}, and ``geometric step decay''~\citep{DavisDC19}. In this paper, we will call it stagewise step decay. The stagewise step decay approach was first introduced in \citep{Goffin77} and used in many \emph{convex} optimization problem~\citep[e.g.,][]{HazanK11,AybatFGO19,KulunchakovM19,GeKKN19}. Interestingly, \citet{GeKKN19} also shows promising empirical results on non-convex functions, but instead of using their proposed decay strategy, they use an exponentially decaying schedule, like the one we analyze here.
The only use of the stagewise step decay for non-convex functions we know are for sharp functions~\citep{DavisDC19} and weakly-quasi-convex functions~\citep{YuanYJY19}. However, they do not show any adaptation property and they still do not consider the exponential step size but its discrete version. As far as we know, we prove the first theoretical guarantee for the exponential step size.

\textbf{Cosine step decay} Cosine step decay was originally presented in~\citet{LoshchilovH17} with two tunable parameters. Later, \citet{HeZZZXL19} proposed a simplified version of it with one parameter. However, there is no theory for this strategy though it is popularly used in the practical world \citep{LiuSY18, ZhangHZZXL19, LawenBPFZ19, ZhangWZZ19, GinsburyCHKLNZJ19, CubukZMVL19,ZhaoJK20,YouLHX20,ChenKNH20,GrillSATRBDABGGPKMV20}. 
As far as we know, we prove the first theoretical guarantee for the cosine step decay and the first ones to hypothesize and prove the adaptation properties of the cosine decay step size.

%\textbf{Line-search for SGD} \citet{VaswaniMLSGLJ19} proposed a line-search technique to set the step sizes. They provide convergence analysis for both convex and non-convex problems. However, they require that the objective function should have a finite-sum structure and satisfy the strong growth condition which implies that we are in the interpolation regime. In contrast, we do not assume a finite-sum structure nor the strong growth condition.
%What we need is a rather general assumption on the stochasticity of the noisy gradients: $\E_t \| \nabla f(\bx_t)- g_t \|^2 \leq a\| \nabla f(\bx_t) \|^2 + b$. It tells that the variance of the noise is upper bounded by the squared gradient norm and an additional constant. This actually covers the expected strong growth condition (by setting $b=0$)~\citep{KhaledR20}.

\textbf{SGD on non-convex smooth functions} The first paper to analyze SGD on smooth functions with generic step sizes is \citet{GhadimiL13}. Their analysis show that the optimal step size strategy strongly depends on the level of noise, but they do not offer any automatic strategy to adapt to it.

\textbf{SGD with the PL condition} %\citet{KarimiNS16} show that for a smooth function, strong convexity and Restricted Secant Inequality (RSI)~\citep{ZhangY13} are both special cases of PL.
%Also, the two forms of one-point convexity summarized in~\citet{Zhu18b} are special cases of RSI and PL, thus in turn imply the PL condition.
The PL condition was proposed by \citet{Polyak63} and \citet{Lojasiewicz63}. It is the weakest assumption we know to prove linear rates on non-convex functions.
For SGD, \citet{KarimiNS16} proved the rate of $O\left(1 / \mu^2 T\right)$ for polynomial step sizes assuming Lipschitz and smooth functions, where $\mu$ is the PL constant. Note that the Lipschitz assumption hides the dependency of convergence and step sizes from the noise. It turns out that the Lipschitz assumption is not necessary to achieve the same rate, see Theorem~\ref{thm:pl_smooth_poly_step} in the Appendix. Considering functions with finite-sum structure, \citet{ReddiHSPS16}, \citet{LeiJCJ17} and \citet{LiBZR20} proved improved rates for variance reduction methods.
The convergence rate that we show for the exponential step size is new in the literature on minimization of PL functions. Independently and the same time\footnote{The first version of \citet{KhaledR20} was released on Feb. 9th 2020 on ArXiv while our very first version was available online on Feb. 12th 2020 on ArXiv as well.} with us, \citet{KhaledR20} obtained the same convergence result in the PL condition for SGD with a stepsize that is constant in the first half and then decreases polynomially.
%However, the result is not true for the classic SGD and it might be important to note that variance reduction methods seem to have problems in deep learning applications~\citep{DefazioB18}.
%As far as we know, the convergence rate that we show for the exponential step size is new in the entire literature on minimization of PL functions.
%As far as we know, $O\left(1 / \mu^2 T\right)$ is the best-known rate for non-convex SGD under the PL condition and we match it in Theorem~\ref{thm:pl_smooth_cst_noise}.

\section{Problem Set-up}
\label{sec:def}

\textbf{Notation}
We denote vectors by bold letters, e.g., $\bx \in \R^d$. We denote by $\E [ \cdot ]$ the expectation with respect to the underlying probability space and by $\E_t [ \cdot ]$ the conditional expectation with respect to the past. Any norm in this work is the $\ell_2$ norm. 

\textbf{Setting and Assumptions}
We consider the unconstrained optimization problem $\min_{ \bx \in \R^d }  \  f(\bx)$, where $f(\bx):\R^d\rightarrow \R$ is a function bounded from below and we denote its infimum by $f^{\star}$. Note that we do \emph{not} require $f$ to be convex \emph{nor} to have a finite-sum structure.

We focus on SGD, where, after an initialization of the first iterate as any $\bx_1 \in \R^d$, in each round $t = 1, 2, \dots, T$ we receive $\bg_t$, an unbiased estimate of the gradient of $f$ at point $\bx_t$, i.e., $\E_t \bg_t = \nabla f(\bx_t)$. We update $\bx_t$ with a step size $\eta_t$, i.e., $\bx_{t+1} = \bx_t - \eta_t \bg_t$.

We assume that 
\begin{itemize}[topsep=0pt]
\item[(\textbf{A1})]
$f$ is \emph{$L$-smooth}, i.e., $f$ is differentiable and its gradient $\nabla f(\cdot)$ is $L$-Lipschitz, namely: $\forall \bx, \by \in \R^d$, $\| \nabla f(\bx) - \nabla f(\by) \| \leq L \| \bx - \by \|$. This implies for $\forall \bx, \by \in \R^d$~\citep[Lemma 1.2.3]{Nesterov04}
{\small{\begin{equation}
\label{eq:smooth}
\left|f(\by)-f(\bx)-\langle \nabla f(\bx), \by-\bx\rangle\right|
\leq \frac{L}{2}\|\by-\bx\|^2~.
\end{equation}}}
\item[(\textbf{A2})] 
$f$ satisfies the \emph{$\mu$-PL} condition, that is, for some $\mu > 0$, $\frac{1}{2} \| \nabla f(\bx) \|^2 \geq \mu \left( f(\bx) - f^{\star} \right), \ \forall \bx$.

%In words, the gradient grows at least as the square root of the sub-optimality.
\item[(\textbf{A3})] For $t = 1, 2, \dots, T$, we assume {\small{$\E_t [\| \bg_t - \nabla f(\bx_t) \|^2 ] \leq a \| \nabla f(\bx_t) \|^2 + b $}}, where $a, b \geq 0$. 

%\citet{BottouCN16} establishes a linear convergence for SGD with convex functions under this assumption.
\end{itemize}

\textbf{Discussion on the assumptions}
It is worth stressing that non-convex functions are not characterized by a particular property, but rather from the lack of a specific property: convexity. In this sense, trying to carry out any meaningful analyses on the entire class of non-convex functions is hopeless. So, the assumptions we use balance the trade-off of \emph{approximately} model many interesting machine learning problems while allowing to restrict the class of non-convex functions on particular subsets where we can underline interesting behaviours.

More in detail, the smoothness assumption (\textbf{A1}) is considered ``weak'' and ubiquitous in analyses of optimization algorithms in the non-convex setting. In many neural networks, it is only approximately true because ReLUs activation functions are non-smooth. However, if the number of training points is large enough, it is a good approximation of the loss landscape.

On the other hand, the PL condition (\textbf{A2}) is often considered a ``strong'' condition. However, it was formally proved to hold locally in deep neural networks in \citet{Allen-ZhuLS19}. Furthermore, \citet{KleinbergLY18} empirically observed that the loss surface of neural networks has good one-point convexity properties, and thus locally satisfies the PL condition. Of course, in our theorems we only need it to hold along the optimization path and not over the entire space, as also pointed out in \citet{KarimiNS16}. So, while being strong, it actually models the cases we are interested in.
Moreover, dictionary learning~\citep{AroraGMM15}, phase retrieval~\citep{ChenC15}, and matrix completion~\citep{SunL16}, all satisfy the one-point convexity locally~\citep{Zhu18b}, and in turn they all satisfy the PL condition locally.

Our assumption on the noise (\textbf{A3}) is strictly weaker than the common assumption of assuming a bounded variance, i.e., {\footnotesize{$\E_t [\| \bg_t - \nabla f(\bx_t) \|^2 ] \leq \sigma^2$}}. Indeed, our assumption recovers the bounded variance case with $a=0$ while also allowing for the variance to grow unboundedly far from the optimum when $a>0$. This is indeed the case when the optimal solution has low training error and the stochastic gradients are generated by mini-batches. This relaxed assumption on the noise was first used by \citet{BertsekasT96} in the analysis of the asymptotic convergence of SGD. 

\textbf{Exponential and Cosine Step Size}
We will use the following definition for the exponential step size
\begin{equation}
\label{eq: step_size}
\eta_t = \eta_0 \cdot \alpha^t
\end{equation}
and for cosine step sizes
\begin{equation}
\label{eq:cosine_step}
\eta_t = \frac{\eta_0 }{2}\left(1+ \cos \frac{t\pi}{T}\right),  
\end{equation}
where $\eta_0 = (L(1+a))^{-1}$. 
For the exponential step sizes, we use $\alpha = \left(\beta/T\right)^{\nicefrac{1}{T}} \leq 1$, $a$ and $L$ are defined in (\textbf{A1}, \textbf{A3}), and $\beta \geq 1$.
%We can see that $\alpha$ goes to 1 when $T$ goes to infinity.

\section{Convergence and Adaptivity of Cosine and Exponential Step Sizes}
\label{sec:theorem}
Here, we present the guarantees of the exponential step size and the cosine step size and their adaptivity property.

\subsection{Noise and Step Sizes}
\label{sec:step}
For the stochastic optimization of smooth functions, the noise plays a crucial role in setting the optimal step sizes: \emph{To achieve the best performance, we need two completely different step size decay schemes in the noisy and noiseless case}. In particular, if the PL condition holds, in the noise-free case a constant step size is used to get a linear rate (i.e., exponential convergence), while in the noisy case the best rate $O(1/T)$ is given by time-varying step sizes $O(1/(\mu t))$~\citep{KarimiNS16}. Similarly, without the PL condition, we still need a constant step size in the noise-free case for the optimal rate whereas a $O(1/\sqrt{t})$ step size is required in the noisy case~\citep{GhadimiL13}.
Using a constant step size in noisy cases is of course possible, but the best guarantee we know is converging towards a neighborhood of the critical point or the optimum, instead of the exact convergence let alone the adaptivity to the noise, as shown in Theorem 2.1 of~\citep{GhadimiL13} and Theorem 4 of~\citep{KarimiNS16}.
Moreover, if the noise decreases over the course of the optimization, we should change the step size as well. Unfortunately, noise levels are rarely known or measured. On the other hand, an optimization algorithm \emph{adaptive to noise} would always get the best performance without changing its hyperparameters.

%\emph{This means that for each noise level, namely mini-batch size in the finite-sum scenario, we need to tune a different step size decay to obtain the best performance.} This process is notoriously tedious and time-consuming.

%Another choice is the stagewise step decay.
%For example, \citet{GeKKN19} propose to start from a constant step size and cut it by a fixed factor every $O(\ln T)$ steps, decaying roughly to $O\left(1/T\right)$ after $T$ iterations.
%However, in practice, deciding when to cut the step size becomes a series of hyperparameters to tune, making this strategy difficult to use in real-world applications. 

%We show that the above problems can be solved by using the exponential step sizes
In the following, we will show that exponential and cosine step sizes achieve exactly this adaptation to noise. It is worth reminding the reader that \emph{any} polynomial decay of the step size does not give us this adaptation. 
So, let's gain some intuition on why this should happen with these two step sizes.
In the early stage of the optimization process, we can expect that the disturbance due to the noise is relatively small compared to how far we are from the optimal solution. Accordingly, at this phase, a near-constant step size should be used. More precisely, the proofs shows that to achieve a linear rate we need $\sum_{t=1}^T \eta_t = \Omega(T)$ or even $\sum_{t=1}^T \eta_t = \Omega(T/ \ln T)$. This is exactly what happens with \eqref{eq: step_size} and \eqref{eq:cosine_step}. On the other hand, when the iterate is close to the optimal solution, we have to decrease the step size to fight with the effects of the noise. In this stage, the exponential step size goes to 0 as $O \left(1/T \right)$, which is the optimal step size used in the noisy case. 
Meanwhile, the last $i$th cosine step size is $\eta_{T-i} = \frac{\eta_0}{2}(1- \cos\frac{i \pi }{T})= \eta_0 \sin^2 \frac{i\pi}{2T}$, which amounts $O (1/T^2)$ when $i$ is much smaller than $T$.

Hence, the analysis shows that \eqref{eq: step_size} and \eqref{eq:cosine_step} are surprisingly similar, smoothly varying from the near-constant behavior at the start and decreasing with a similar pattern towards the end, and both will be adaptive to the noise level.
%Particularly, the exponential step size is emulating the transition between the \emph{optimal} constant one at the beginning and \emph{optimal} decreasing one towards the end in a smooth continuous way.
Next, we formalize these intuitions in convergence rates.

\subsection{Convergence Guarantees}
\label{ssec:guarantee}

We now prove the convergence guarantees for these two step sizes.
First, we consider the case where the function is smooth and satisfies the PL condition.
\begin{theorem}[SGD with exponential step size]
\label{thm: pl_smooth_cst_noise}
Assume (\textbf{A1}, \textbf{A2}, \textbf{A3}). For a given $T \geq \max\{3, \beta \}$ and $\eta_0 = (L(1+a))^{-1}$, with step size \eqref{eq: step_size}, SGD guarantees 
\begin{align*}
& \E f(\bx_{T+1}) - f^{\star} \leq \frac{5LC(\beta)}{e^2 \mu^2 } \frac{\ln^2 \frac{T}{\beta}}{T} b\\
& \quad + C(\beta) \exp\left(-\frac{0.69\mu }{L+a} \left(\frac{T}{\ln \frac{ T}{\beta}}\right)\right)\cdot (f(\bx_1) - f^{\star}), 
\end{align*}
where $C(\beta)\triangleq \exp \left((2\mu\beta)/(L (1+a)\ln T/\beta)\right)$.
\end{theorem}

\textbf{Choice of $\beta$} 
Note that if $\beta = L(1+a)/\mu$, we get 
\begin{align*}
& \E f(\bx_{T+1}) - f^{\star} \\
& \leq O\left(\exp\left(-\frac{\mu }{L+a} \left(\frac{T}{\ln \frac{\mu T}{L}}\right)\right)+ \frac{b \ln^2 \frac{\mu T}{L}}{\mu^2  T} \right)~.
\end{align*}
In words, this means that we are basically free to choose $\beta$, but will pay an exponential factor in the mismatch between $\beta$ and $\frac{L}{\mu}$, which is basically the condition number for PL functions. This has to be expected because it also happens in the easier case of stochastic optimization of strongly convex functions~\citep{MoulinesB11}.
\begin{theorem}[SGD with cosine step size]
\label{thm:PL_cosine}
Assume (\textbf{A1}, \textbf{A2}, \textbf{A3}). For a given $T$ and $\eta_0 = (L(1+a))^{-1}$, with step size \eqref{eq:cosine_step}, SGD guarantees 
\begin{align*}
& \E f(\bx_{t+1}) - f^{\star} 
\leq \exp \left(- \frac{\mu (T-1)}{2L(1+a)}\right) (f(x_1) - f^{\star})\\
& \quad + \frac{ \pi^4 b}{32 (1+a)T^4} \left( \left(\frac{8T^2}{\mu}\right)^{4/3} + \left(\frac{6T^2}{\mu}\right)^{\frac{5}{3}}\right) ~.
\end{align*}
\end{theorem}

\textbf{Adaptivity to Noise} From the above theorems, we can see that both the exponential step size and the cosine step size have a provable advantage over polynomial ones: \emph{adaptivity to the noise}. Indeed, when $b=0$, namely there is only noise relative to the distance from the optimum, they both guarantee a linear rate. Meanwhile, if there is noise, using the \emph{same step size without any tuning}, the exponential step size recovers the rate of $O\left(1/(\mu^2 T)\right)$ while the cosine step size achieves the rate of $O(1/(\mu^{\frac{5}{3}}T^{\frac{2}{3}}))$ (up to poly-logarithmic terms). In contrast, polynomial step sizes would require two different settings---decaying vs constant---in the noisy vs no-noise situation~\citep{KarimiNS16}.
It is worth stressing that the rate in Theorem~\ref{thm: pl_smooth_cst_noise} is one of the first results in the literature on stochastic optimization of smooth PL functions \citep{KhaledR20}.
%It is possible to obtain similar but incomparable rates for strongly convex and smooth functions~\citep{Zhu18,AybatFGO19}, but here we do not need convexity.

\textbf{Optimality of the bounds} As far as we know, it is unknown if the rate we obtain for the optimization of non-convex smooth functions under the PL condition is optimal or not. However, up to poly-logarithmic terms, Theorem~\ref{thm: pl_smooth_cst_noise} matches at the same time the best-known rates for the noisy and deterministic cases~\citep{KarimiNS16} (see also Theorem~\ref{thm:pl_smooth_poly_step} in the Appendix). We would remind the reader that this rate is not comparable with the one for strongly convex functions which is $O(1/(\mu T))$.
Meanwhile, cosine step size achieves a rate slightly worse in $T$ (but better in $\mu$) under the same assumptions.

\textbf{Cosine Step Size with Restarts} The original cosine stepsize was proposed with a restarting strategy, yet it has been commonly used without restarting and achieves good results~\citep[e.g.,][]{LoshchilovH17,Gastaldi17,ZophVSL18,HeZZZXL19,CubukZMVL19,LiuSY18,ZhaoJK20,YouLHX20,ChenKNH20,GrillSATRBDABGGPKMV20}. Indeed, the previous theorem has confirmed that the cosine stepsize alone is well worth studying theoretically. Yet for completeness, we cover the analysis in a restart scheme for SGD with cosine stepsize in the PL condition in Appendix~\ref{ssec:restart}. We obtain the same convergence rate $\mu $ and $T$ as that in the case of no restarts under the PL condition.

\textbf{Convergence without the PL condition} The PL condition tells us that all stationary points are optimal points~\citep{KarimiNS16}, which is not always true for the parameter space in deep learning~\citep{JinGNKJ17}. However, this condition might still hold locally, for a considerable area around the local minimum. Indeed, as we said, this is exactly what was proven for deep neural networks~\citep{Allen-ZhuLS19}.
The previous theorems tell us that once we reach the area where the geometry of the objective function satisfies the PL condition, we can get to the optimal point with an almost linear rate, depending on the noise. Nevertheless, we still have to be able to reach that region.
Hence, in the following, we discuss the case where the PL condition is not satisfied and show for both step sizes that they are still able to move to a critical point at the optimal speed.
\begin{theorem}
\label{thm:no_PL_no_noise}
Assume \textbf{(A1)}, \textbf{(A3)} and $c>1$. SGD with step sizes \eqref{eq: step_size} with $\eta_0 = (c L(1+a))^{-1}$ guarantees
\begin{align*}
\E  \| \nabla f(\tilde{\bx}_T) \|^2
& \leq \frac{3 L c (a+1)\ln \frac{T}{\beta}}{T- \beta} \cdot \left(f(\bx_1) - f^{\star}\right)\\
& \quad + \frac{b T}{c(a+1)(T-\beta)}, 
\end{align*}
where $\tilde{\bx}_T$ is a random iterate drawn from $\bx_1, \dots, \bx_T$ with $\Pr[\tilde{\bx}_T=\bx_t]= \frac{\eta_t}{\sum_{i=1}^T \eta_i }$.
\end{theorem}
\begin{theorem}
\label{thm:no_PL_cosine}
Assume \textbf{(A1)}, \textbf{(A3)} and $c>1$. SGD with step sizes \eqref{eq:cosine_step} with $\eta_0 = (c L(1+a))^{-1}$ guarantees
\begin{align*}
\E  \| \nabla f(\tilde{\bx}_T) \|^2
& \leq  \frac{4 L c (a+1)}{T- 1} \cdot \left(f(\bx_1) - f^{\star}\right)\\
& \quad + \frac{21 b T}{4 \pi^4 c L (a+1)(T-1)}, 
\end{align*}
where $\tilde{\bx}_T$ is a random iterate drawn from $\bx_1, \dots, \bx_T$ with $\Pr[\tilde{\bx}_T=\bx_t]= \frac{\eta_t}{\sum_{i=1}^T \eta_i }$.
\end{theorem}
If $b\neq 0$ in \textbf{(A3)}, setting $c \propto \sqrt{T}$ and $\beta=O(1)$ would give the $\tilde{O}(1/\sqrt{T})$ rate\footnote{The $\tilde{O}$ notations hides poly-logarithmic terms.} and $O(1/\sqrt{T})$ for the exponential and cosine step size respectively. Note that the optimal rate in this setting is $O(1/\sqrt{T})$. On the other hand, if $b=0$, setting $c=O(1)$ and $\beta=O(1)$ yields a $\tilde{O}(1/T)$ rate and $O(1/T)$ for the exponential and cosine step size respectively. It is worth noting that the condition $b=0$ holds in many practical scenarios~\citep{VaswaniBS19}.
Note that both guarantees are optimal up to poly-logarithmic terms~\citep{ArjevaniCDFSW19}.

In the following, we present the main elements of the proofs of these theorems, leaving the technical details in the Appendix. The proofs also show the mathematical similarities between these two step sizes.

\textbf{Proofs of the Theorems}
Given that the space is limited, we defer the proofs of Theorem~\ref{thm:no_PL_no_noise} and Theorem~\ref{thm:no_PL_cosine} to the Appendix.

We first introduce some technical lemmas whose proofs are in the Appendix.
\begin{lemma}
\label{lemma:start}
Assume \textbf{(A1)}, \textbf{(A3)}, and $\eta_t \leq \frac{1}{L(1+a)}$.  SGD guarantees 
\begin{equation}
\label{eq:thm2_eq1}
\begin{split}
\E f(\bx_{t+1})- \E  f(\bx_t)
\leq -  \frac{\eta_t}{2} \E \| \nabla f(\bx_t) \|^2 + \frac{L \eta_t^2 b}{2}~. 
\end{split}
\end{equation}
\end{lemma}
\begin{lemma}
\label{lemma: ratio_bound}
Assume $X_k, A_k, B_k \geq 0, k = 1 ,...$, and $X_{k+1} \leq A_k X_k + B_k$, then we have 
\[
X_{k+1} \leq \prod_{i=1}^k A_i X_1 + \sum_{i=1}^{k} \prod_{j=i+1}^k A_j B_i~. 
\]
\end{lemma}
\begin{lemma}
\label{lemma:sum_cosine}
For $\forall T \geq 1$, we have $\sum_{t=1}^{T} \cos\frac{t \pi}{T} = -1$.
\end{lemma}
\begin{lemma}
\label{lemma: ineq_constant}
For $T \geq 3$, $\alpha \geq 0.69$ and $\frac{ \alpha^{T+1}}{(1-\alpha)} \leq \frac{2\beta}{\ln \frac{T}{\beta}}$. 
\end{lemma}
\begin{lemma}
\label{lemma: ineq_alpha}
$
1-x \leq \ln \left(\frac{1}{x}\right), \forall x > 0. 
$
\end{lemma}
\begin{lemma}
\label{lemma: integral_bound}
Let $a,b\geq0$. Then 
\[
\sum_{t=0}^T \exp(-b t) t^a \leq 2\exp(-a)\left(\frac{a}{b}\right)^a+ \frac{\Gamma(a+1)}{b^{a+1}}~.
\]
\end{lemma}
We can now prove both Theorem~\ref{thm: pl_smooth_cst_noise} and Theorem~\ref{thm:PL_cosine}.
\begin{proof}[Proof of Theorem~\ref{thm: pl_smooth_cst_noise} and Theorem~\ref{thm:PL_cosine}.]
Denote $\E f(\bx_t) - f^{\star}$ by $\Delta_t$. From Lemma~\ref*{lemma:start} and the PL condition, we get
\begin{equation}
\Delta_{t+1} 
\leq  (1 - \mu \eta_t ) \Delta_t + \frac{L}{2} \eta_t^2 b^2~. 
\end{equation} 
By Lemma~\ref{lemma: ratio_bound} and $1- x \leq \exp (-x)$, we have 
\begin{align}
& \Delta_{T+1} 
  \leq \prod_{t=1}^{T} (1- \mu \eta_t) \Delta_1 +  \frac{L}{2} \sum_{t=1}^{T} \prod_{i=t+1}^{T} (1- \mu \eta_i) \eta_t^2 b\\
&  \leq \exp \left(- \mu \sum_{t=1}^{T} \eta_t \right) \Delta_1  +  \frac{Lb}{2} \sum_{t=1}^{T} \exp \left( - \mu \sum_{i=t+1}^{T} \eta_i \right) \eta_t^2. 
\end{align}
We then show that both the exponential step size and the cosine step size satisfy $\sum_{t=1}^{T} \eta_t = \Omega (T)$, which guarantees a linear rate in the noiseless case.

For the cosine step size \eqref{eq:cosine_step}, we observe that 
\begin{align*}
\sum_{t=1}^{T} \eta_t
%&  = \sum_{t=1}^{T} \frac{\eta_0 }{2}(1+ \cos \frac{t\pi}{T}) \\
 = \frac{\eta_0 T}{2} + \frac{\eta_0}{2} \sum_{t=1}^{T} \cos \frac{t\pi}{T} = \frac{\eta_0 (T-1)}{2},
\end{align*}
where in the last equality we used Lemma~\ref{lemma:sum_cosine}. 

Also, for the exponential step size \eqref{eq: step_size}, we can show
\begin{align*}
 \sum_{t=1}^{T} \eta_t = \eta_0 \frac{\alpha - \alpha^{T+1}}{1- \alpha}
 & \geq  \frac{\eta_0 \alpha}{1- \alpha} - \frac{2\eta_0\beta }{\ln \frac{T}{\beta}}\\
 & \geq  T \cdot  \frac{0.69\eta_0}{\ln \frac{T}{\beta}} -  \frac{2\eta_0\beta }{\ln \frac{T}{\beta}}, 
\end{align*}
where we used Lemma~\ref{lemma: ineq_constant} in the first inequality and Lemma~\ref{lemma: ineq_alpha} in the second inequality. 

Next, we upper bound $\sum_{t=1}^{T} \exp \left( - \mu \sum_{i=t+1}^{T} \eta_i \right) \eta_t^2$ for these two kinds of step sizes respectively.

For the exponential step size, by Lemma~\ref{lemma: ineq_constant}, we obtain
\begin{align*}
& \sum_{t=1}^{T} \exp \left( - \mu \sum_{i=t+1}^{T} \eta_i \right) \eta_t^2 \\
& = \eta_0^2 \sum_{t=1}^{T} \exp \left( - \mu \eta_0 \frac{\alpha^{t+1} - \alpha^{T+1}}{1- \alpha}\right) \alpha^{2t}\\
& \leq \eta_0^2 C(\beta) \sum_{t=1}^{T} \exp \left( - \frac{ \mu \eta_0\alpha^{t+1}}{1- \alpha}\right) \alpha^{2t}\\
& \leq \eta_0^2 C(\beta)  \sum_{t=1}^{T} \left(\frac{e}{2} \frac{\mu \alpha^{t+1}}{L(1+a)(1-\alpha)}\right)^{-2} \alpha^{2t}\\
& \leq \frac{4L^2(1+a)^2}{e^2\mu^2} \sum_{t=1}^T\frac{1}{\alpha^2} \ln^2 \left(\frac{1}{\alpha}\right)
\leq  \frac{10 L^2(1+a)^2\ln^2 \frac{T}{\beta}}{e^2 \mu^2 T}, 
\end{align*}
where in the second inequality we used $\exp(-x) \leq \left(\frac{\gamma}{e x}\right)^\gamma, \forall x >0, \gamma>0$. 

For the cosine step size, using the fact that $\sin x \geq \frac{2}{\pi}x$ for $ 0 \leq x \leq \frac{\pi}{2}$, we can lower bound $\sum_{i=t+1}^{T} \eta_i $ by
\begin{align*}
\sum_{i=t+1}^{T} \eta_i 
&= \frac{\eta_0 }{2}\sum_{i=t+1}^{T}  \left(1+ \cos \frac{i \pi}{T}\right) \\
%&= \frac{\eta_0 }{2}\sum_{i=0}^{T-t-1}  \left(1- \cos \frac{i \pi}{T}\right) \\
& = \frac{\eta_0 }{2}\sum_{i=0}^{T-t-1}  \sin^2 \frac{i \pi}{2T} 
 \geq \frac{\eta_0}{2T^2}\sum_{i=0}^{T-t-1}  i^2 \\
&\geq \frac{\eta_0(T-t-1)^3}{6T^2}~. 
\end{align*}
Then, we proceed 
\begin{align*}
& \sum_{t=1}^{T} \exp \left(- \mu \sum_{i=t+1}^{T} \eta_i\right) \eta_t^2 \\
& \leq \frac{\eta_0^2}{4} \sum_{t=1}^{T} \left(1+ \cos \frac{t\pi}{T}\right)^2 \exp \left(- \frac{\mu \eta_0(T-t-1)^3}{6T^2}\right)\\
% & = \frac{ \eta_0^2}{4} \sum_{t=0}^{T-1} (1+ \cos \frac{(T-t)\pi}{T})^2 \exp \left(- \frac{\eta_0 \mu (t-1)^3}{6T^2}\right) \\
& = \frac{ \eta_0^2}{4} \sum_{t=1}^{T-1} \left(1-  \cos \frac{t\pi}{T}\right)^2 \exp \left(- \frac{\eta_0 \mu (t-1)^3}{6T^2}\right) \\
& =\eta_0^2\sum_{t=1}^{T-1} \sin^4 \frac{t\pi}{2T} \exp \left(- \frac{\eta_0 \mu (t-1)^3}{6T^2}\right) \\
& \leq \frac{\eta_0^2 \pi^4}{16T^4}\sum_{t=0}^{T-1}t^4  \exp \left(- \frac{\eta_0 \mu t^3}{6T^2}\right)\\
& \leq \frac{\eta_0 \pi^4}{16 T^4} \left(2 \exp\left(-\frac{4}{3}\right) \left(\frac{8T^2}{\mu}\right)^{4/3} + \left(\frac{6T^2}{\mu}\right)^{\frac{5}{3}}\right), 
\end{align*}
where in the third line we used $\cos(\pi-x) = - \cos(x)$, in the forth line we used $1- \cos(2x) = 2\sin^2(x)$, and in the last inequality we applied Lemma~\ref{lemma: integral_bound}. 

Putting things together, we get the stated bounds. 
\end{proof}

\begin{figure*}[t]
\centering
\includegraphics[width=\textwidth]{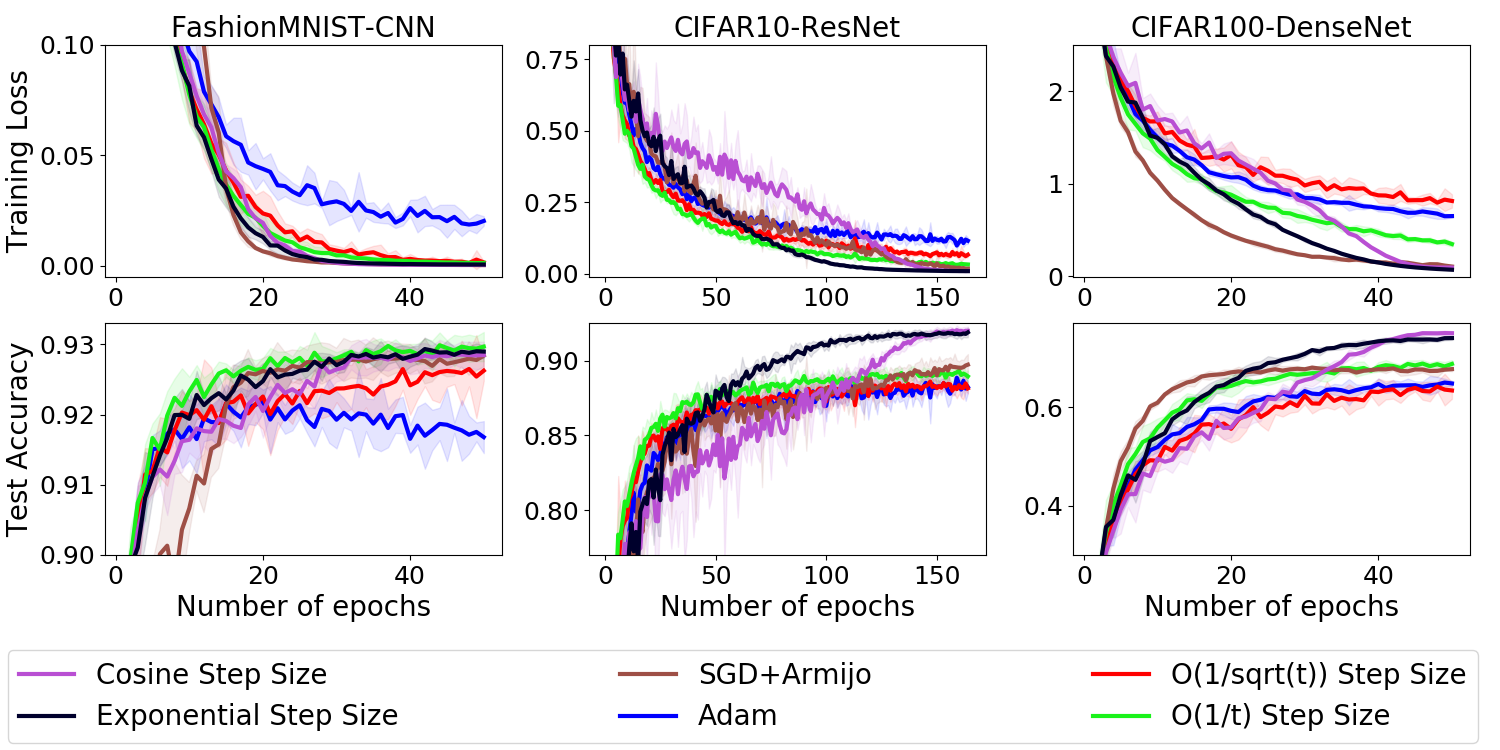}
\vspace{-2em}
\caption{Training loss (top plots) and test accuracy (bottom plots) curves on employing different step size schedules to do image classification using a simple CNN for FashionMNIST (left), a 20-layer ResNet for CIFAR-10 (middle), and a 100-layer DenseNet on CIFAR-100 (right). \emph{(The shading of each curve represents the 95\% confidence interval computed
across five independent runs from random initial starting points.)}}%\vspace{-1em}
\label{fig:stepsize}
\end{figure*}

\section{Empirical Results}
\label{sec:exp}

The empirical performance of the cosine step size is already well-known in the applied world and does not require additional validation. However, both the exponential and the cosine step size are often missing as baselines in recent empirical evaluations. Hence, the main aim of this section is to provide a comparison of the exponential and cosine step sizes to other popular state-of-the-art step sizes methods. All experiments are done in PyTorch~\cite{Pytorch19} and the codes can be found at \url{https://github.com/zhenxun-zhuang/SGD-Exponential-Cosine-Stepsize}.

We performed experiments using deep neural networks to do image classification tasks on various datasets with different network architectures. Additionally, Appendix A.3.3 features an experiment on a Natural Language Processing (NLP) task, where the exponential and cosine step size strategies obtain better results than Adam~\citep{KingmaB15}, the de-facto optimization method in NLP. Finally, in Appendix A.3.1, we include a synthetic experiment where those assumptions we need in analysis hold and show in detail the noise adaptation of both step sizes as predicted by the theory.

All models and experiments were carefully chosen to be easily reproducible.

\textbf{Datasets}
We consider the image classification task on FashionMNIST and CIFAR-10/100 datasets. For all datasets, we select 10\% training images as the validation set. Data augmentation and normalization are described in the Appendix.

\textbf{Models}
For FashionMNIST, we use a CNN model consisting of two alternating stages of $5\times5$ convolutional filters and $2\times2$ max-pooling followed by one fully connected layer of 1024 units. To reduce overfitting, 50\% dropout noise is used during training. For the CIFAR-10 dataset, we employ the 20-layer Residual Network model~\citep{HeZRS16}; and for CIFAR-100, we utilize the DenseNet-BC model~\citep{HuangLVW17} with 100 layers and a growth rate of 12. The loss is cross-entropy. The codes for implementing the latter two models can be found here\footnote{\url{https://github.com/akamaster/pytorch_resnet_cifar10}} and here\footnote{\url{https://github.com/bearpaw/pytorch-classification}} respectively.

\begin{figure*}[t]
\centering
\includegraphics[width=\textwidth]{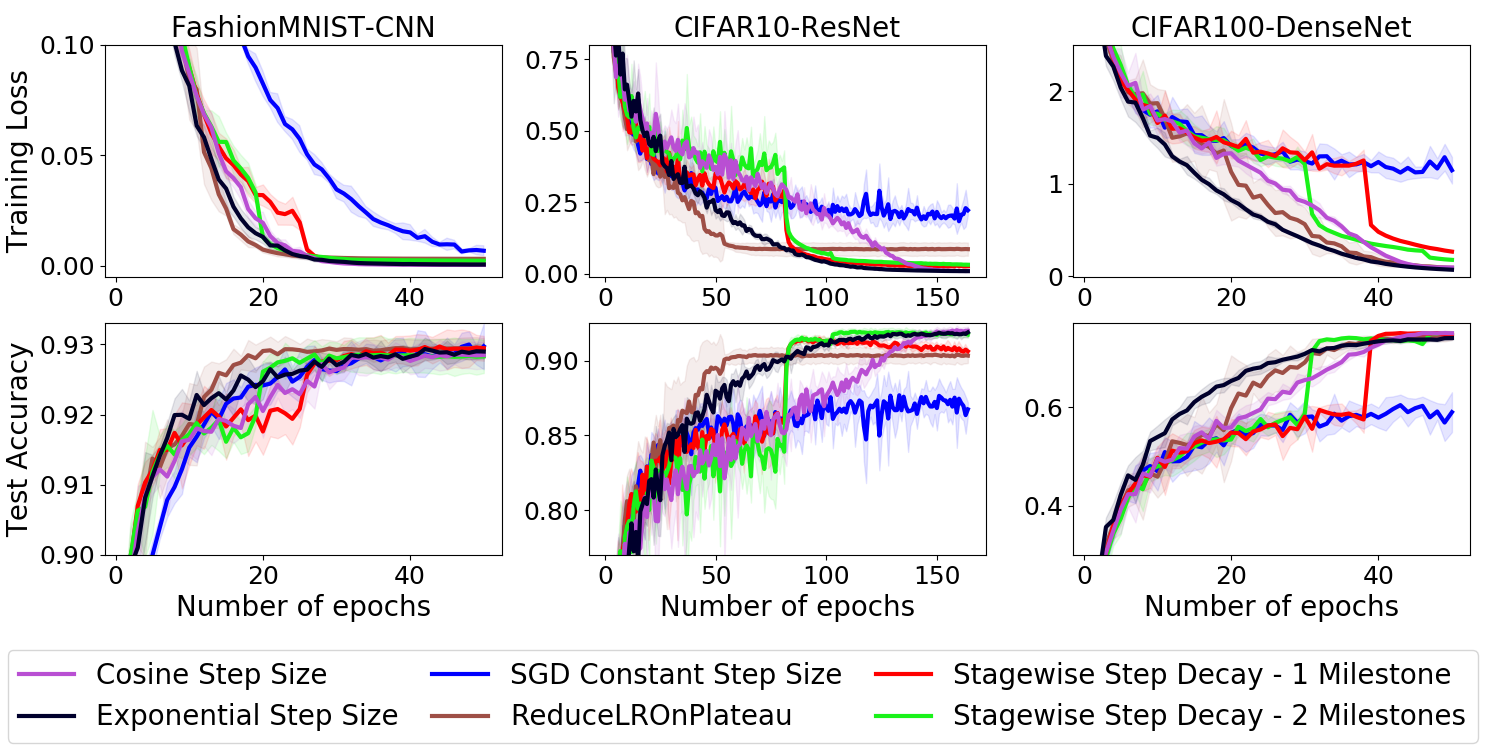}
\vspace{-2em}
\caption{Training loss (top plots) and test accuracy (bottom plots) curves comparing the exponential and cosine step sizes with stagewise step decay for image classification using a simple CNN for FashionMNIST (left), a 20-layer ResNet for CIFAR-10 (middle), and a 100-layer DenseNet on CIFAR-100 (right). \emph{(The shading of each curve represents the 95\% confidence interval computed
across five independent runs from random initial starting points.)}}%\vspace{-12pt}
\label{fig:stagewise}
\end{figure*}

\textbf{Training}
During the validation stage, we tune each method using the grid search (full details in the Appendix) to select the hyperparameters that work best according to their respective performance on the validation set. At the testing stage, the best performing hyperparameters from the validation stage are employed to train the model over all training images. The testing stage is repeated with random seeds for 5 times to eliminate the influence of stochasticity.

We use Nesterov momentum~\citep{Nesterov83} of 0.9 without dampening (if having this option), weight-decay of 0.0001 (FashionMNIST and CIFAR-10) and 0.0005 (CIFAR100), and use a batch size of 128. Regarding the employment of Nesterov momentum, we follow the setting of~\citet{GeKKN19}. The use of momentum is essential to have a fair and realistic comparison in that the majority of practitioners would use it when using SGD. 
%It is a plus instead of a minus that these two step sizes couple well with these techniques.

\textbf{Optimization methods} We consider SGD with the following step size decay schedules:
\begin{equation}\label{eq:decays}
\begin{split}
&\eta_t = \eta_0\cdot\alpha^t; \quad
\eta_t = \eta_0(1+\alpha\sqrt{t})^{-1}; \\
&\eta_t = \eta_0(1+\alpha t)^{-1}; \quad
\eta_t = \eta_0/2 \left(1+\cos\left(t\pi/T\right)\right),
\end{split}
\end{equation}
where $t$ is the iteration number (instead of the number of epochs). We also compare with Adam~\citep{KingmaB15}, SGD+Armijo~\citep{VaswaniMLSGLJ19}, PyTorch's ReduceLROnPlateau scheduler\footnote{\url{https://pytorch.org/docs/stable/optim.html}} and stagewise step decay. In the following, we will call the place of decreasing the step size in stagewise step decay a \textbf{milestone}. (As a side note, since we use Nesterov momentum in all SGD variants, the stagewise step decay basically covers the performance of multistage accelerated algorithms \citep[e.g.,][]{AybatFGO19}.)

\begin{table*}[h!]
\caption{Average final training loss and test accuracy achieved by each method when optimizing respective models on each dataset. The $\pm$ shows $95\%$ confidence intervals of the mean loss/accuracy value over 5 runs starting from different random seeds.}
\label{tab:results}
{\tiny
\begin{tabular}{|c|c|c|c|c|c|c|}
\hline
\multirow{2}{*}{Methods} & \multicolumn{2}{c|}{FashionMNIST} & \multicolumn{2}{c|}{CIFAR10} & \multicolumn{2}{c|}{CIFAR100} \\
\cline{2-7}
& Training loss & Test accuracy & Training loss & Test accuracy & Training loss & Test accuracy \\
\hline
SGD Constant Step Size & $0.0068 \pm 0.0023$ & $0.9297 \pm 0.0033$ & $0.2226 \pm 0.0169$ & $0.8674 \pm 0.0048$ & $1.1467 \pm 0.1437$ & $0.5896 \pm 0.0404$ \\
\hline
$O(1/t)$ Step Size & $0.0013 \pm 0.0004$ & $\mathbf{0.9297 \pm 0.0021}$ & $0.0331 \pm 0.0028$ & $0.8894 \pm 0.0040$ & $0.3489 \pm 0.0263$ & $0.6874 \pm 0.0076$ \\
\hline
$O(1/\sqrt{t})$ Step Size & $0.0016 \pm 0.0005$ & $0.9262 \pm 0.0014$ & $0.0672 \pm 0.0086$ & $0.8814 \pm 0.0034$ & $0.8147 \pm 0.0717$ & $0.6336 \pm 0.0169$ \\
\hline
Adam & $0.0203 \pm 0.0021$ & $0.9168 \pm 0.0023$ & $0.1161 \pm 0.0111$ & $0.8823 \pm 0.0041$ & $0.6513 \pm 0.0154$ & $0.6478 \pm 0.0054$ \\
\hline
SGD+Armijo & $\mathbf{0.0003 \pm 0.0000}$ & $0.9284 \pm 0.0016$ & $0.0185 \pm 0.0043$ & $0.8973 \pm 0.0071$ & $0.1063 \pm 0.0153$ & $0.6768 \pm 0.0044$ \\
\hline
ReduceLROnPlateau & $0.0031 \pm 0.0009$ & $0.9294 \pm 0.0015$ & $0.0867 \pm 0.0230$ & $0.9033 \pm 0.0049$ & $0.0927 \pm 0.0085$ & $0.7435 \pm 0.0076$ \\
\hline
Stagewise - 1 Milestone & $0.0007 \pm 0.0002$ & $0.9294 \pm 0.0018$ & $0.0269 \pm 0.0017$ & $0.9062 \pm 0.0020$ & $0.2673 \pm 0.0084$ & $0.7459 \pm 0.0030$ \\
\hline
Stagewise - 2 Milestones & $0.0023 \pm 0.0005$ & $0.9283 \pm 0.0024$ & $0.0322 \pm 0.0008$ & $0.9174 \pm 0.0020$ & $0.1783 \pm 0.0030$ & $0.7487 \pm 0.0025$ \\
\hline
Exponential Step Size & $0.0006 \pm 0.0001$ & $0.9290 \pm 0.0009$ & $\mathbf{0.0098 \pm 0.0010}$ & $0.9188 \pm 0.0033$ & $\mathbf{0.0714 \pm 0.0041}$ & $0.7398 \pm 0.0037$ \\
\hline
Cosine Step Size & $0.0004 \pm 0.0000$ & $0.9285 \pm 0.0019$ & $0.0106 \pm 0.0008$ & $\mathbf{0.9199 \pm 0.0029}$ & $0.0949 \pm 0.0053$ & $\mathbf{0.7497 \pm 0.0044}$ \\
\hline
\end{tabular}}
\end{table*}

\textbf{Results and discussions}
The exact loss and accuracy values are reported in Table~\ref{tab:results}. To avoid overcrowding the figures, we compare the algorithms in groups of baselines. The comparison of performance between step size schemes listed in~\eqref{eq:decays}, Adam, and SGD+Armijo are shown in Figure~\ref{fig:stepsize}. As can be seen, the \emph{only} two methods that perform well on \emph{all} 3 datasets are cosine and exponential step size. In particular, cosine step size performs the best across datasets both in training loss and test accuracy, with the exponential step size following closely.

On the other hand, as we noted above, stagewise step decay is a very popular decay schedule in deep learning. Thus, our second group of baselines in Figure~\ref{fig:stagewise} is composed by the stagewise step decay, ReduceLROnPlateau, and SGD with constant stepsize. The results show that exponential and cosine step sizes can still match or exceed the best of them with a fraction of their needed time to find the best hyperparameters. Indeed, we need 4 hyperparameters for two milestones, 3 for one milestone, and at least 4 for ReduceLROnPlateau. In contrast, the cosine step size requires only 1 hyperparameter and the exponential one needs 2.

Note that we do not pretend that our benchmark of the stagewise step decay is exhaustive. Indeed, there are many unexplored (potentially infinite!) possible hyperparameter settings. For example, it is reasonable to expect that adding even more milestones at the appropriate times could lead to better performance.
However, this would result in a linear growth of the number of hyperparameters leading to an exponential increase in the number of possible location combinations.
%we can see that, for stagewise step decay, the first milestone witnesses a dramatic improvement in both the training loss and the test accuracy curves, and the second milestone contributes further improvements.
%Obviously, these milestones where one shrinks the step size are the key factor of the good performance of stagewise step decay.
\begin{wrapfigure}{c}{0.6\linewidth}
\begin{center}
\includegraphics[width=\linewidth]{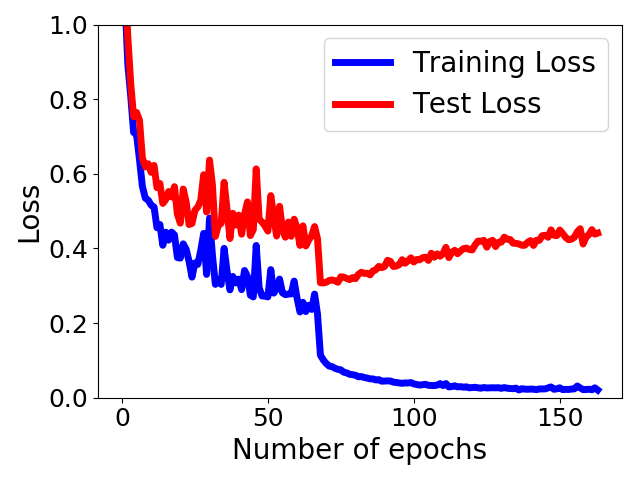}
\end{center}
\caption{Plot showing that decreasing the step size too soon would lead to overfitting (ResNet20 on CIFAR10).}
\label{fig:overfitting}
\end{wrapfigure}
This in turn causes the rapid growth of tuning time in selecting a good set of milestones in practice. Worse still, even the intuition that one should decrease the step size once the test loss curve stops decreasing is not always correct. Indeed, we observed in experiments (see Figure~\ref{fig:overfitting}) that doing this will, after the initial drop of the curve in response to the step size decrease, make the test loss curve gradually go up again.

\section{Conclusion}
\label{sec:disc}

We have analyzed theoretically and empirically the exponential and cosine step size, two successful step size decay schedules for the stochastic optimization of non-convex functions. We have shown that, up to poly-logarithmic terms, both step sizes guarantee convergence with the best-known rates for smooth non-convex functions. Moreover, in the case of functions satisfying the PL condition, we have also proved that they are both adaptive to the level of noise. Furthermore, we have validated our theoretical findings on both synthetic and real-world tasks, showing that these two step sizes consistently match or outperform other strategies, while at the same time requiring only 1 (cosine) / 2 (exponential) hyperparameters to tune.

In future work, we plan to extend our theoretical results. For example, high probability bounds are easy to be obtained from our results.

\section*{Acknowledgements}
This material is based upon work supported by the National Science Foundation under grants no. 1925930 ``Collaborative Research: TRIPODS Institute for Optimization and Learning'', no. 1908111 ``AF: Small: Collaborative Research: New Representations for Learning Algorithms and Secure Computation'', and no. 2022446 ``Foundations of Data Science Institute''.

\balance
\bibliography{learning}
\bibliographystyle{icml2021}
\newpage
\appendix
\onecolumn
\section{Appendix}

\subsection{Convergence for non-Lipschitz PL functions}
\citet{KarimiNS16} proved that SGD with an appropriate step size will give a $O(1/T)$ convergence for Lipschitz and PL functions. However, it is easy to see that the Lipschitz assumption can be substituted by the smoothness one and obtain a rate that depends on the variance of the noise. Even if this is a straightforward result, we could not find it anywhere so we report here our proof.

\begin{theorem}
\label{thm:pl_smooth_poly_step}
Assume \textbf{(A1)} and \textbf{(A3)} and set the step sizes $\eta_t = \min \left(\frac{1}{L(1+a)}, \frac{2t+1}{\mu (t+1)^2}\right)$. Then, SGD guarantees
\begin{align*}
f(\bx_{T+1}) - f^{\star}
\leq \frac{L^2 (1+a)b}{2\mu^3 T^2}+ \frac{2L}{\mu^2 T}b
+ (f(\bx_1) - f^{\star})\frac{L^2(1+a)^2}{\mu^2 T^2} \left(1-\frac{\mu}{L(1+a)}\right)^{\frac{L(1+a)}{\mu}}~.
\end{align*}
\end{theorem}
\begin{proof}
For simplicity, denote $\E f(\bx_t) - f^{\star}$ by $\Delta_t$. With the same analysis as in Theorem~\ref{thm: pl_smooth_cst_noise}, we have
\[
\Delta_{t+1} \leq \left(1- \mu \eta_t\right) \Delta_t + \frac{L}{2} \eta_t^2 b~.
\]
Denote by $t^{\star} = \min \left\{t: \frac{t^2}{2t+1} \leq \frac{L(1+a) - \mu}{\mu}\right\}$. When $t \leq t^{\star}$, $\eta_t = \frac{1}{L(1+a)}$ and we obtain
\[
\Delta_{t+1} \leq \left(1-\frac{\mu}{L(1+a)}\right) \Delta_t + \frac{b}{2L(1+a)^2}~.
\]
Thus, by Lemma \ref{lemma: ratio_bound}, we get
\begin{align*}
\Delta_{t^{\star}}
& \leq \left(1-\frac{\mu}{L(1+a)}\right)^{t^{\star}-1} \Delta_1
+ \frac{b}{2L(1+a)^2}\sum_{i=0}^{t^{\star}}\left(1-\frac{\mu}{L(1+a)}\right)^{t^{\star}-i} \\
& \leq \left(1-\frac{\mu}{L(1+a)}\right)^{t^{\star}} \Delta_1 + \frac{b}{2\mu(1+a)}~.
\end{align*}
Instead, when $t \geq t^{\star}$, $\eta_t = \frac{2t+1}{\mu (t+1)^2}$, we have
\[
\Delta_{t+1} \leq \frac{t^2}{(t+1)^2} \Delta_t + \frac{L (2t+1)^2}{2 \mu^2 (t+1)^4 }b.
\]
Multiplying both sides by $(t+1)^2$ and denoting by $\delta_t = t^2 \Delta_t$, we get
\[
\delta_{t+1} \leq \delta_t + \frac{L(2t+1)^2}{2\mu^2(t+1)^2}b
\leq \delta_t + \frac{2L}{\mu^2}b~.
\]
Summing over $t$ from $t^{\star}$ to $T$, we have
\[
\delta_{T+1} \leq \delta_{t^{\star}\\
} + \frac{2L(T-t^{\star})}{\mu^2} b~.
\]
Then, we finally get
\begin{align*}
\Delta_{T+1}
& \leq \frac{t^{\star2}}{T^2} \left(1-\frac{\mu}{L(1+a)}\right)^{t^{\star}}\Delta_1
+ \frac{t^{\star2}b}{2\mu(1+a)T^2}+ \frac{2L(T-t^{\star})}{\mu^2 T^2}b \\
& \leq \frac{L^2(1+a)^2}{\mu^2 T^2} \left(1-\frac{\mu}{L(1+a)}\right)^{\frac{L(1+a)}{\mu}}\Delta_1
+ \frac{L^2 (1+a)b}{2\mu^3 T^2}+ \frac{2L}{\mu^2 T}b~. \qedhere
\end{align*}
\end{proof}

\subsection{Cosine stepsize with Restarts}
\label{ssec:restart}

Cosine stepsize is proposed with warm restarts~\citep{LoshchilovH17}.  We then complete the theory by providing an analysis of SGD with cosine stepsize in this restarting scheme (Algorithm~\ref{alg:restart}) under the PL condition. The proof builds on the fact that the suboptimality gap shrinks after each restarting and the rate depends on the suboptimality gap at the beginning of each restarting.

\begin{algorithm}
\caption{SGD with Cosine Stepsize and Restarts}
\label{alg:restart}
\begin{algorithmic}
\STATE \textbf{Input:} Initial Step size $\eta_0$, time increase factor $r$, initial point $\bx_1$. 
\FOR{$i = 0, \dots, l$}
\STATE   Let $T_i = T_0 \, r^i $
\FOR{$t = 0, \dots, T_i -1$}
\STATE  Run SGD with cosine stepsize  $\frac{\eta_0}{2} \left(1+ \cos \frac{t \pi}{T_i}\right)$
\ENDFOR
\ENDFOR
\end{algorithmic}
\end{algorithm}
\begin{theorem}[SGD with cosine step size and restart]
\label{thm:PL_cosine_restart}
Assume (\textbf{A1}, \textbf{A2}, \textbf{A3}). For a given $T_0 $, $r >1$ , $T_i = T_0 \, r^i $, $T \triangleq \sum_{i=0}^{l} T_i$, and $\eta_0 = (L(1+a))^{-1}$, Algorithm~\ref{alg:restart} guarantees (where $\tilde{O}$ hides the $log$ terms) 
\begin{align*}
\E f(\bx_T) - f^{\star} 
\leq 
\tilde{O} \left(\exp \left(- \frac{\mu (T - l - 1)}{2L(1+a)} \right) + b\left( \frac{1}{\mu^{4/3} T^{4/3}} + \frac{1}{\mu^{5/3} T^{2/3}}\right) \right),
\end{align*}
and for $r = 1$, it guarantees 
\[
\E f(\bx_T) - f^{\star} 
\leq  C_1 \left(\frac{1}{\mu^{4/3} T_0^{4/3}} + \frac{1}{\mu^{5/3} T_0^{2/3}}\right)  \frac{1- \exp \left(-C_2 \mu(T-l-1)\right)}{1- \exp\left(-C_2 \mu(T_0-1)\right)}+ \exp \left(- \mu C_2 (T-l-1)\right) (f(x_1) - f^{\star}),
\]
where $C_1 \triangleq  \frac{  6^{5/3}\pi^4 b}{32 (1+a)}  $ and $C_2 \triangleq \frac{1}{2L(1+a)}$. 
\end{theorem}
\begin{proof}[Proof of Theorem~\ref{thm:PL_cosine_restart}] Denote by $S_i =\sum_{j=0}^{i}T_j$ and $S_{-1} = 1$. Given Theorem~\ref{thm:PL_cosine},  it is immediate to have $\forall i = 0, \dots, l$:
\begin{align*}
\E f(\bx_{S_i}) - f^{\star} 
& \leq \frac{ \pi^4 b}{32 (1+a)T_i^4} \left( \left(\frac{8T_i^2}{\mu}\right)^{4/3} + \left(\frac{6T_i^2}{\mu}\right)^{\frac{5}{3}}\right) + \exp \left(- \frac{\mu (T_i-1)}{2L(1+a)}\right) (f(x_{S_{i-1}})- f^{\star})\\
& \leq C_1 \left(\frac{1}{\mu^{4/3} T_i^{4/3}} + \frac{1}{\mu^{5/3} T_i^{2/3}}\right) + \exp \left(-  C_2 \mu (T_i-1)\right) (f(x_{S_{i-1}}) - f^{\star})~.
\end{align*}

Repeatedly using the above inequality, we get 
\begin{align*}
\E f(\bx_{S_l}) - f^{\star}  
\leq  C_1 \sum_{i=0}^{l}\prod_{j= i+1}^{l} \exp(- C_2 \mu (T_j - 1))   \left(\frac{1}{\mu^{4/3} T_i^{4/3}} + \frac{1}{\mu^{5/3} T_i^{2/3}}\right) 
+ \exp \left(- C_2\mu  (S_l-l-1)\right) (f(x_1) - f^{\star})~.
\end{align*}
	
In the case of $r= 1$, $T_i = T_0$, we have for any $i$ that
\begin{align*}
\sum_{i=0}^{l}\prod_{j= i+1}^{l} \exp(- C_2 \mu (T_j - 1)) \left(\frac{1}{\mu^{4/3} T_i^{4/3}} + \frac{1}{\mu^{5/3} T_i^{2/3}}\right)
& = \left(\frac{1}{\mu^{4/3} T_0^{4/3}} + \frac{1}{\mu^{5/3} T_0^{2/3}}\right) \sum_{i=0}^{l}\exp \left(- C_2 \mu (T_0 -1) (l-i)\right)   \\
& = \left(\frac{1}{\mu^{4/3} T_0^{4/3}} + \frac{1}{\mu^{5/3} T_0^{2/3}}\right)  \frac{1- \exp \left(-C_2 \mu(T_0 - 1) (l+1)\right)}{1- \exp\left(-C_2 \mu(T_0-1)\right)}~.
\end{align*}

In the case of $r > 1$, denote by $A_i = \prod_{j= i+1}^{l} \exp(- C_2 \mu (T_j - 1))   \frac{1}{\mu^pT_i^{q}} , p,q>0$.  For any $i = 0, ... , l$, $\frac{A_i}{A_{i+1}} = \exp(-C_2 \mu (T_{i+1} -1)) r^q$ is decreasing over $i$. Denote by $i^{\star} = \min \{ i: \frac{A_i}{A_{i+1}} \leq 1\}$.
We have 
\begin{align*}
\sum_{i=0}^{l}A_i 
& \leq A_0 \cdot  i^{\star} + (l - i^{\star} + 1) \cdot A_l \leq  (l + 1) \cdot (A_0 + A_l) 
 =\frac{ l + 1  }{\mu^p}\cdot \left(\frac{1}{T_l^q} + \frac{1}{T_0^q}\exp(- C_2\mu (T-T_0-l))\right), \quad p, q >0.  
\end{align*}
Note that  $T_l = T \cdot \frac{r^l (r-1)}{r^{l+1}-1} \geq \frac{r-1}{r} T$ and $l = O \left(\ln T \right)$. Then, the stated bound follows. 
\end{proof}

\subsection{Proofs in Section~\ref{sec:theorem}}
\begin{proof}[Proof of Lemma~\ref{lemma:start}]
By \eqref{eq:smooth}, we have
\begin{equation}
\label{eq:smooth_one_step}
f(\bx_{t+1}) \leq f(\bx_t) - \langle \nabla f(\bx_t), \eta_t\bg_t \rangle + \frac{L}{2} \eta_t^2 \| \bg_t \|^2~.
\end{equation}
Taking expectation on both sides, we get
\begin{align*}
\E f(\bx_{t+1})- \E f(\bx_t)
\leq - \left(\eta_t - \frac{L(a+1)}{2} \eta_t^2 \right)\E \| \nabla f(\bx_t) \|^2 + \frac{L}{2}\eta_t^2 b 
\leq - \frac{1}{2}\eta_t \E \| \nabla f(\bx_t) \|^2 + \frac{L}{2}\eta_t^2 b,
\end{align*}
where in the last inequality we used the fact that $\eta_t \leq \frac{1}{L(1+a)}$.
\end{proof}
\begin{proof}[Proof of Lemma~\ref{lemma: ratio_bound}]
When $k=1$, $X_2 \leq A_1 X_1 + B_1$ satisfies. By induction, assume $X_{k} \leq \prod_{i=1}^{k-1} A_i X_1 + \sum_{i=1}^{k-1}\prod_{j=i+1}^{k-1} A_j B_i$, and we have
\begin{align*}
X_{k+1}
&\leq A_k \left( \prod_{i=1}^{k-1} A_i X_1 + \sum_{i=1}^{k-1} \prod_{j=i+1}^{k-1} A_j B_i \right) + B_k 
= \prod_{i=1}^k A_i X_1 + \sum_{i=1}^{k-1} \prod_{j=i+1}^{k} A_j B_i + A_k B_k \\
&= \prod_{i=1}^k A_i X_1 + \sum_{i=1}^k \prod_{j=i+1}^k A_j B_i~. \qedhere
\end{align*}
\end{proof}
\begin{proof}[Proof of Lemma~\ref{lemma: ineq_constant}]
We have
\begin{align*}
\frac{\alpha^{T+1}}{(1-\alpha)}
= \frac{\alpha \beta }{T (1-\alpha)}
= \frac{\beta }{T\left(1 - \exp\left(-\frac{1}{T} \ln \frac{T}{\beta}\right)\right)}
\leq \frac{2\beta }{\ln \frac{T}{\beta}},
\end{align*}
where in the last inequality we used $\exp(-x) \leq 1- \frac{x}{2}$ for $0 < x < \frac{1}{e}$ and the fact that $\frac{1}{T} \ln\left(\frac{T}{\beta}\right) \leq \frac{\ln T}{T} \leq \frac{1}{e}$.
\end{proof}
\begin{proof}[Proof of Lemma~\ref{lemma:sum_cosine}]
If $T$ is odd, we have
\begin{align*}
\sum_{t=1}^{T} \cos\frac{t \pi}{T}
= \cos \frac{T\pi }{T} + \sum_{t=1}^{(T-1)/2} \cos \frac{t \pi}{T} + \cos \frac{(T-t)\pi}{T} 
 = \cos \pi = -1,
\end{align*}
where in the second inequality we used the fact that $\cos (\pi - x) = - \cos (x)$ for any $x$.
If $T$ is even, we have
\[
\sum_{t=1}^{T} \cos\frac{t \pi}{T}
= \cos \frac{T\pi }{T} + \cos \frac{T\pi }{2T} + \sum_{t=1}^{T/2 - 1} \cos \frac{t \pi}{T} + \cos \frac{(T-t)\pi}{T} 
= \cos \pi = -1~. \qedhere
\]
\end{proof}
\begin{proof}[Proof of Lemma~\ref{lemma: ineq_alpha}]
It is enough to prove that $f(x) := x - 1- \ln x \geq 0$. Observe that $f'(x)$ is increasing and $f'(1) = 0$, hence, we have $f(x) \geq f(1) = 0$.
\end{proof}
\begin{proof}[Proof of Lemma~\ref{lemma: integral_bound}]
Note that $f(t)=\exp(-b t) t^a$ is increasing for $t\in [0,a/b]$ and decreasing for $t\geq a/b$. Hence, we have
\begin{align*}
\sum_{t=0}^T \exp(-b t) t^a
&\leq \sum_{t=0}^{\lfloor a/b \rfloor-1} \exp(-b t) t^a + \exp(-b \lfloor a/b \rfloor) \lfloor a/b \rfloor^a + \exp(-b \lceil a/b \rceil) \lceil a/b \rceil^a +\sum_{\lceil a/b \rceil+1}^T \exp(-b t) t^a \\
&\leq 2\exp(-a)(a/b)^a+\int_{0}^{\lfloor a/b \rfloor} \exp(-b t) t^a dt + \int_{\lceil a/b \rceil}^T \exp(-b t) t^a dt\\
&\leq 2\exp(-a)(a/b)^a+\int_{0}^{T} \exp(-b t) t^a dt \\
&\leq 2\exp(-a)(a/b)^a+\int_{0}^{\infty} \exp(-b t) t^a dt \\
&=2\exp(-a)(a/b)^a+ \frac{1}{b^{a+1}} \Gamma(a+1)~. \qedhere
\end{align*}
\end{proof}

\begin{proof}[Proof of Theorem \ref{thm:no_PL_no_noise} and Theorem \ref{thm:no_PL_cosine}]
We observe that for exponential step sizes,
\begin{align*}
\sum_{t=1}^T \eta^2_t
\leq \frac{\alpha^2}{L^2 c^2 (a+1)^2(1- \alpha^2)}~.
\end{align*}
and for cosine step sizes,
\begin{align*}
\sum_{t=1}^{T} \eta_t^2
= \frac{\eta_0^2}{4} \sum_{t=1}^{T} \left(1+ \cos\frac{t\pi}{T}\right)^2 = \frac{\eta_0^2}{4} \sum_{t=0}^{T-1} \left(1- \cos\frac{t\pi}{T}\right)^2 
= \eta_0^2 \sum_{t=1}^{T} \sin^4\frac{t\pi}{2T} \leq \eta_0^2 \sum_{t=1}^{T} \frac{t^4\pi^4}{16T^4}
%= \frac{\eta_0^2 (T+1)(2T+1) (3T^2+3T+1) }{16T^3 \pi^4}
\leq \frac{21 \eta_0^2 T}{8 \pi^4}~.
\end{align*}
Summing \eqref{eq:thm2_eq1} over $t=1,\dots,T$ and dividing both sides by $\sum_{t=1}^T \eta_t$, we get the stated bound.
\end{proof}

\subsection{Experiments details}
\label{ssec:expdet}

\subsubsection{Image classification experiments.}
\paragraph{Data Normalization and Augmentation.}
Images are normalized per channel using the means and standard deviations computed from all training images. For CIFAR-10/100, we adopt the data augmentation technique following~\citet{LeeXGZT15} (for training only): 4 pixels are padded on each side of an image and a $32\times32$ crop is randomly sampled from the padded image or its horizontal flip.

\paragraph{Hyperparameter tuning.} We tune the hyperparameters on the validation set using the following two-stage grid searching strategy. First, search over a coarse grid, and select the one yielding the best validation results. Next, continue searching in a fine grid centering at the best performing hyperparameters found in the coarse stage, and in turn, take the best one as the final choice.

For the starting step size $\eta_0$, the coarse searching grid is \{0.00001, 0.0001, 0.001, 0.01, 0.1, 1\}, and the fine grid is like \{0.006, 0.008, 0.01, 0.02, 0.04\} if the best one in the coarse stage is 0.01.

For the $\alpha$ value, we set its searching grid so that the ratio $\eta_T/\eta_0$, where $\eta_T$ is the step size in the last iteration, is first searched over the coarse grid of \{0.00001, 0.0001, 0.001, 0.01, 0.1, 1\}, and then over a fine grid centered at the best one of the coarse stage. Note that we try all pairs of $(\eta_0, \alpha)$ from their respective searching grids.

For the stagewise step decay, to make the tuning process more thorough, we modify as follows the one employed in Section 6.1 (specifically on tuning SGD V1) of \citet{YuanYJY19}, where they first set two milestones and then tune the starting step size. Put it explicitly and take the experiment on CIFAR-10 as an example, we first run vanilla SGD with a constant step size to search for a good range of starting step size on the grid \{0.00001, 0.0001, 0.001, 0.01, 0.1, 1\}, and find 0.01 and 0.1 work well. Based on this, we set the fine searching grid of starting step sizes as \{0.007, 0.01, 0.04, 0.07, 0.1, 0.4\}. For each of them, we run three settings with an increasing number of milestones: vanilla SGD (with no milestone), SGD with 1 milestone, and SGD with 2 milestones. The searching grid for milestones is \{16k, 24k, 32k, 40k, 48k, 56k\} (number of iterations). For the 1 milestone setting, the milestone can be any of them. For the 2 milestones, they can be any combination of two different elements from the searching grid, like (16k, 32k) or (32k, 48k). The grid search strategy for FashionMNIST and CIFAR-100 is similar but with the searching grid for milestones over \{3k, 6k, 9k, 12k, 15k, 18k\}.

The PyTorch ReduceLROnPlateau scheduler takes multiple arguments, among which we tune the starting learning rate, the factor argument which decides by which the learning rate will be reduced, the patience argument which controls the number of epochs with no improvement after which learning rate will be reduced, and the threshold argument which measures the new optimum to only focus on significant changes. We choose the searching grid for the starting step size using the same strategy for stagewise step decay above, i.e., first running SGD with a constant step size to search for a good starting step size, then search over a grid centering on the found value, which results in the grid \{0.004, 0.007, 0.01, 0.04, 0.07\} (FashionMNIST) and \{0.01, 0.04, 0.07, 0.1, 0.4\} (CIFAR10/100). We also explore the searching grid of the factor argument over \{0.1, 0.5\}, the patience argument over \{5, 10\} (CIFAR10) or \{3, 6\} (FashionMNIST/CIFAR100), and the threshold argument over \{0.0001, 0.001, 0.01, 0.1\}.

For each setting, we choose the combination of hyperparameters that gives the best final validation loss to be used in testing. Also, whenever the best performing hyperparameters lie in the boundary of the searching grid, we always extend the grid to make the final best-performing hyperparameters fall into the interior of the grid.

\subsubsection{Natural language inference}
\paragraph{Dataset} We conduct this experiment on the Stanford Natural Language Inference (SNLI) dataset~\citep{BrownAPM15} which contains 570k pairs of human-generated English sentences. Each pair of sentences is manually labeled with one of three categories: entailment, contradiction, and neutral, and thus forms a three-way classification problem. It captures the task of natural language inference, a.k.a.~Recognizing Textual Entailment (RTE).

\paragraph{Model} We employ the bi-directional LSTM of about 47M parameters proposed by ~\citet{ConneauKSBB17}. Except for replacing the cross-entropy loss with an SVM loss following~\citet{BerradaZK19}, we leave all other components unchanged (codes can be found here\footnote{\url{https://github.com/oval-group/dfw}}). Like them, we also use the open-source GloVe vectors~\citep{PenningtonSM14} trained on Common Crawl 840B with 300 dimensions as fixed word embeddings.

\paragraph{Training} During the validation stage, we tune each method using the grid search. The initial learning rate of each method is grid searched over $\{0.00001, 0.0001, 0.001, 0.01, 0.1, 1, 10\}$. And the $\alpha$ of the exponential step size is searched over a grid such that the ratio $\eta_T/\eta_0$, where $\eta_T$ is the step size in the last iteration, is over \{0.0001, 0.001, 0.01, 0.1, 1\}. Following~\citep{BerradaZK19}, for each hyperparameter setting, we record the best validation accuracy obtained during training and select the setting that performs the best according to this metric to do the test. The testing stage is repeated with different random seeds for 5 times to eliminate the influence of stochasticity.

We employ the Nesterov momentum~\citep{Nesterov83} of 0.9 without dampening (if having this option), but do not use weight decay. The mini-batch size is 64 and we run for 10 epochs.

\begin{figure*}[t]
\centering
\includegraphics[width=\textwidth]{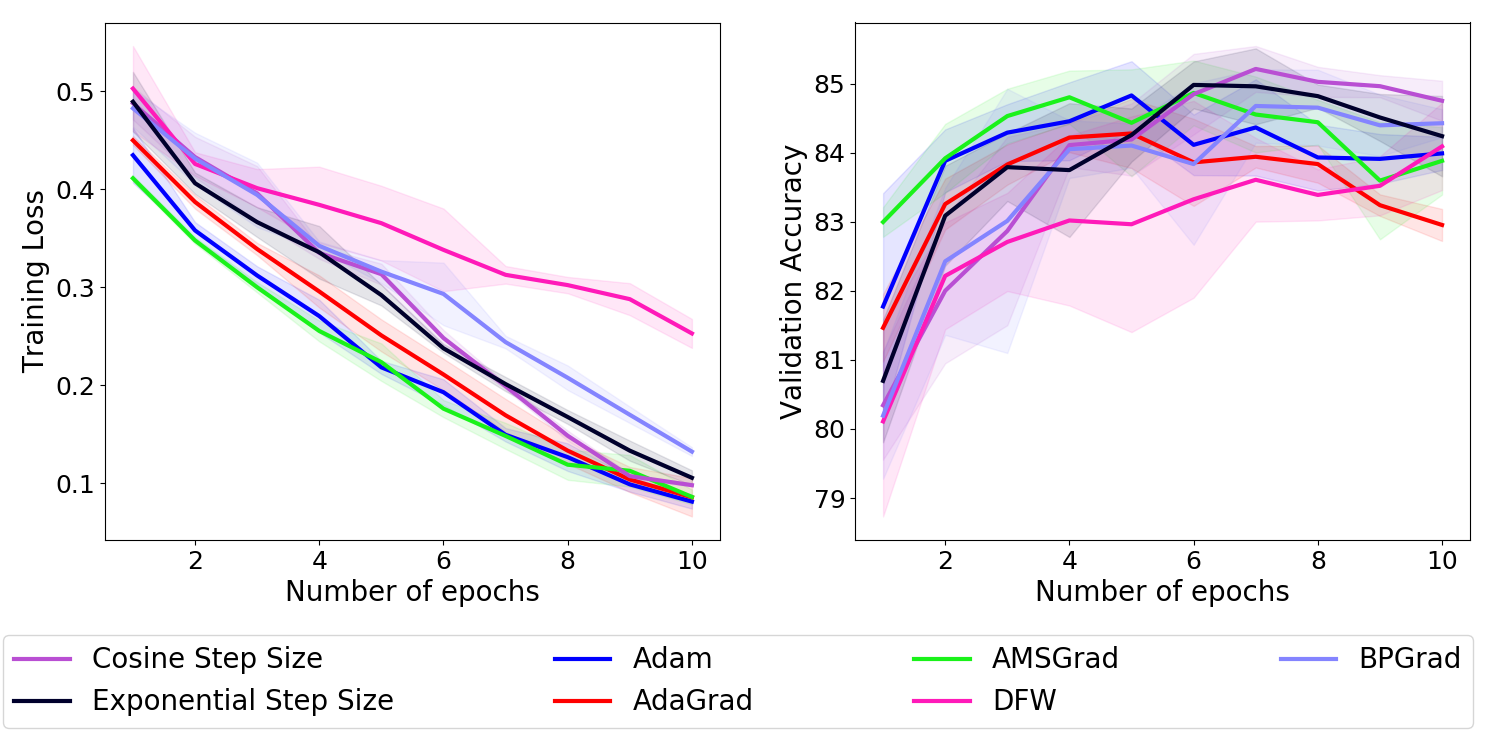}
\caption{Training loss and validation accuracy curves, averaged over 5 independent runs, on using different methods to optimize a Bi-LSTM to do natural language inference on the SNLI dataset. \emph{(The shading of each curve represents the 95\% confidence interval computed
across five independent runs from random initial starting points.)}}
\label{fig:nlp}
\end{figure*}

\paragraph{Results} We compare the exponential and the cosine step sizes with Adagrad, Adam, AMSGrad~\citep{ReddiKK18}, BPGrad~\citep{ZhangWW18}, and DFW~\citep{BerradaZK19}. From Figure~\ref{fig:nlp} and Table~\ref{tab:nlp}, we can see that cosine step size remains the best among all methods, with exponential step size following closely next.

\begin{table}[h]
\caption{The best test accuracy achieved by each method. The $\pm$ shows $95\%$ confidence intervals of the mean accuracy value over 5 runs starting from different random seeds.}
\label{tab:nlp}
\centering
\begin{tabular}{|c|c|}
\hline
Methods & Test Accuracy\\
\hline
Adam & 0.8479 $\pm$ 0.0043\\
\hline
AdaGrad & 0.8446 $\pm$ 0.0027\\
\hline
AMSGrad & 0.8475 $\pm$ 0.0029\\
\hline
DFW & 0.8412 $\pm$ 0.0045\\
\hline
BPGrad & 0.8459 $\pm$ 0.0030\\
\hline
Exp. Step Size & 0.8502 $\pm$ 0.0028\\
\hline
Cosine Step Size & \textbf{0.8509} $\pm$ \textbf{0.0033}\\
\hline
\end{tabular}
\end{table}

\subsubsection{Synthetic Experiments}
We have theoretically proved that exponential and cosine step size can adapt to the level of noise automatically and converge to the optimum without the need to re-tune the hyperparameters. In contrast, for other optimization methods, re-tuning is typically critical for convergence. To validate this empirically, we conduct an experiment on a non-convex function $g(r, \theta) = (2 + \frac{\cos \theta}{2} + \cos 4\theta) r^2 (5/3 - r)$~\citep{ZhouMBBG17}, where $r$ and $\theta$ are the polar coordinates, which satisfies the PL condition when $r\le 1$.

\paragraph{Proof of PL condition.}
We now prove that $f(x, y) = g(r, \theta) = (2 + \frac{\cos \theta}{2} + \cos 4\theta) r^2 (5/3 - r)$ satisfies the PL condition when $r \leq 1$.

Obviously, $2 + \frac{\cos \theta}{2} + \cos 4\theta \ge \frac12$ as $\cos\theta\in[-1, 1]$.

When $r \leq 1$, $\frac53 - r\geq \frac23$, thus $f(x, y)\ge0$, and $f^{\star} = f(0, 0) = 0$.

We first calculate derivatives in polar coordinates
\begin{align*}
\frac{\partial g}{\partial r} &= \left( \frac{10r}{3}- 3r^2 \right) \left(2+ \frac{\cos \theta}{2} + \cos 4\theta\right),\\
\frac{\partial g}{\partial \theta} &= \left(- \frac{\sin \theta}{2} - 4 \sin 4\theta\right) r^2 \left(\frac{5}{3} - r\right)~.
\end{align*}

Then, from the relationship between derivatives in Cartesian and polar coordinates, we have
\begin{align*}
\frac{\| \nabla f(x, y)\|^2}{2(f(x, y) - f^{\star})}
& = \frac{ \left(\frac{\partial g}{\partial r}\right)^2 + \frac{1}{r^2}\left(\frac{\partial g}{\partial \theta}\right)^2 }{2(2+ \frac{\cos \theta }{2} + \cos 4\theta) r^2 (\frac{5}{3} - r)}\\
& = \frac{(\frac{10}{3} - 3r)^2(2+ \frac{\cos \theta}{2} + \cos 4\theta)}{\frac{10}{3} - 2r}
+ \frac{(- \frac{\sin \theta}{2} - 4 \sin 4\theta)^2 (\frac{5}{3} - r) }{2(2+\frac{\cos \theta}{2}+ \cos 4\theta)}\\
& \geq \frac{(\frac{10}{3}- 3r)^2}{4(\frac{5}{3}- r)} \geq \frac{1}{24}~.
\end{align*}

We compare SGD with decay rules listed in~\eqref{eq:decays}, SGD with constant step size, and Adam on optimizing this function. We consider three cases: the noiseless case where we get the exact gradient in each round, the slightly noisy case in which we add independent Gaussian noise with zero mean and standard deviation 0.05 to each dimension of the gradient in each round, and the noisy case with additive Gaussian noise of standard deviation 1.

\begin{figure*}[t]
\centering
\includegraphics[width=\textwidth]{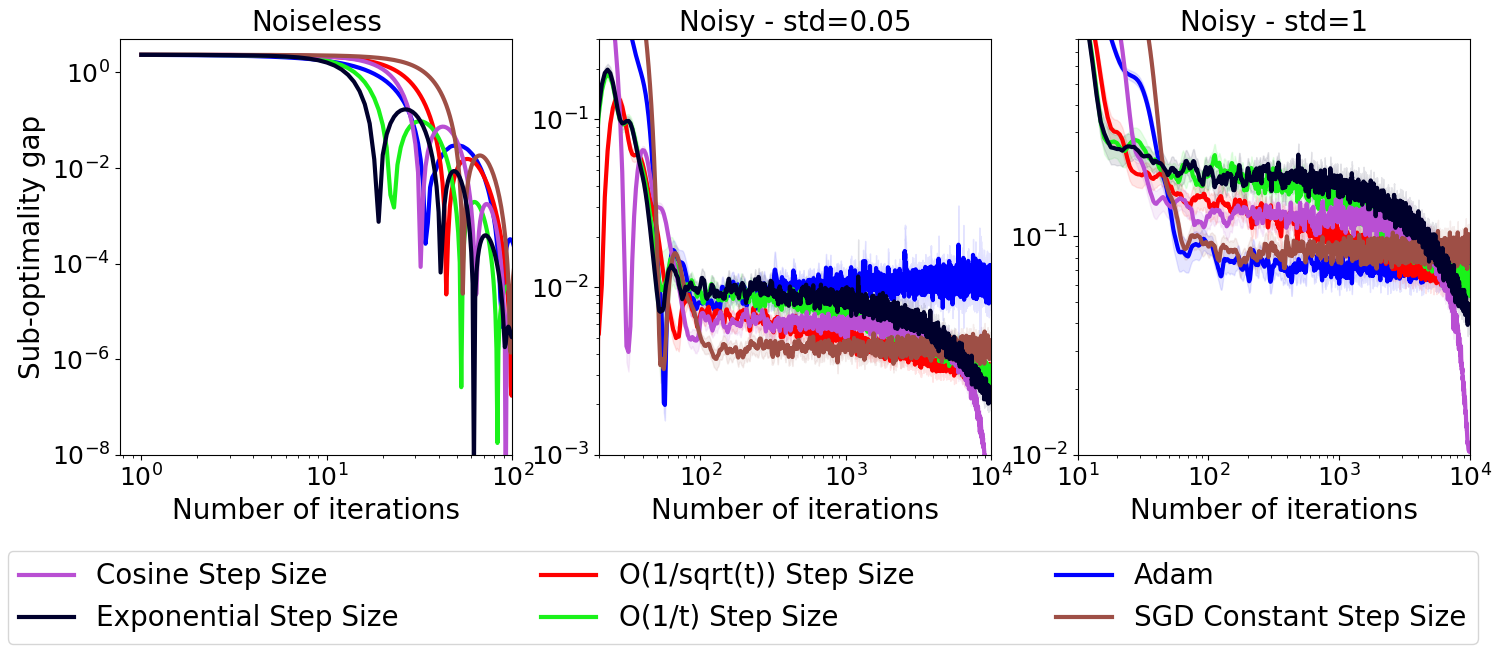}
\caption{Plots of the sub-optimality gap vs. iterations for optimizing a synthetic function. Both axes in all figures are on the logarithmic scale. The left plot is the noiseless case, the middle one is with the additive Gaussian noise of standard deviation 0.05, while the right plot is with the additive Gaussian noise of standard deviation 1. \emph{(The shading of each curve represents the 95\% confidence interval computed
across five independent runs from random initial starting points.)}}
\label{fig:synthetic}
\end{figure*}

We tune hyperparameters for all methods on the noiseless case such that they all obtain roughly the same performance after 100 iterations. We then apply those methods directly to the two noisy cases using the same set of hyperparameters obtained in the noiseless case. To reduce the influence of stochasticity, we average on 100 independent runs with different random seeds. Results shown in Figure~\ref{fig:synthetic} demonstrate that both exponential and cosine step size behave as the theory predicts and has no problem on converging towards the optimum when the noise level changes. In contrast, after the initial decrease in the sub-optimality gap, other methods all end up oscillating around some value which is method-specific and related to the noise level.

\end{document}